\documentclass[11pt]{article}

\usepackage[preprint]{acl}

\usepackage{times}
\usepackage{latexsym}
\usepackage[T1]{fontenc}
\usepackage[utf8]{inputenc}
\usepackage{microtype}
\usepackage{inconsolata}
\usepackage{graphicx}
\usepackage{amsmath}
\usepackage{amssymb}
\usepackage{booktabs}
\usepackage{multirow}
\usepackage{pgfplots}
\usepackage{enumitem}
\usepackage{xspace}
\pgfplotsset{compat=1.18}

\definecolor{amethyst}{rgb}{0.6, 0.4, 0.8}

\definecolor{lemon}{RGB}{255,247,0}
\definecolor{maize}{RGB}{250,237,94}
\definecolor{mustard}{RGB}{255,219,89}
\definecolor{ocre}{RGB}{241,103,35}
\definecolor{Tangerine}{RGB}{253,128,8}
\definecolor{framegreen}{RGB}{153, 188, 133}
\definecolor{bggreen}{RGB}{235, 250, 228}
\definecolor{c0}{cmyk}{1,0.3968,0,0.2588} 
\definecolor{c1}{cmyk}{0,0.6175,0.8848,0.1490} 
\definecolor{c2}{cmyk}{0.1127,0.6690,0,0.4431} 
\definecolor{c3}{cmyk}{0.3081,0,0.7209,0.3255} 
\definecolor{c4}{RGB}{164, 16, 52}
\definecolor{orange}{HTML}{E66100}
\definecolor{bluex}{HTML}{0C7BDC}
\definecolor{yellow}{HTML}{FFC20A}
\definecolor{lightpurple}{HTML}{E6E6FA}
\definecolor{lightbluee}{HTML}{e8f4f8}
\definecolor{blush}{rgb}{0.87, 0.36, 0.51}
\definecolor{c5}{HTML}{EE4E4E}
\definecolor{gggggg}{HTML}{EFEFEF}
\ifnum1=1

\usepackage{inconsolata}
\usepackage[T1]{fontenc}
\usepackage[scaled=0.95]{biolinum}

\usepackage{tcolorbox}
\definecolor{chart}{HTML}{1f77b4}

\newtcolorbox{example}[1][]{
  colback=chart!5!white,
  colframe=chart,
  floatplacement=floating,
  title=\centering \textsf{\small #1}
}

\newtcbox{\hlprimarytab}{on line, box align=base, colback=BlueGreen!20,colframe=blue,size=fbox,arc=3pt, before upper=\strut, top=-2.5pt, bottom=-4.5pt, left=-2pt, right=-2pt, boxrule=0pt}
\newtcbox{\hlsecondarytab}{on line, box align=base, colback=WildStrawberry!10,colframe=orange,size=fbox,arc=3pt, before upper=\strut, top=-2.5pt, bottom=-4.5pt, left=-2pt, right=-2pt, boxrule=0pt}
\newtcbox{\hlwhite}{on line, box align=base, colback=WildStrawberry!8,colframe=white,size=fbox,arc=2pt, before upper=\strut, top=-3pt, bottom=-4.5pt, left=-2pt, right=-2pt, boxrule=0pt}
\newtcbox{\hlyellow}{on line, box align=base, colback=BlueGreen!10,colframe=white,size=fbox,arc=2pt, before upper=\strut, top=-3pt, bottom=-4.5pt, left=-2pt, right=-2pt, boxrule=0pt}

\setlist[itemize]{
    leftmargin=1.0em,  
    rightmargin=0pt, 
    itemsep=0.25em,     
    topsep=0.5em       
}

\newcommand{\thebench}{{RWKU+}\xspace}

\title{On the Hidden Costs of Counterfactual \\Knowledge Training in LLM Unlearning}

\author{%
Xiaotian Ye\textsuperscript{1}\thanks{Equal contribution.},
Xiaohan Wang\textsuperscript{2}$^*$,
Mengqi Zhang\textsuperscript{3},
Shu Wu\textsuperscript{4} 
\\
\textsuperscript{1}Beijing University of Posts and Telecommunications\\
\textsuperscript{2}Huazhong University of Science and Technology\quad
\textsuperscript{3}Shandong University\\
\textsuperscript{4}NLPR, MAIS, Institute of Automation, Chinese Academy of Sciences
\\
\texttt{yexiaotian@bupt.edu.cn, shawn\_wang@hust.edu.cn}\\
\texttt{mengqi.zhang@sdu.edu.cn, shu.wu@nlpr.ia.ac.cn}
}

\begin{document}
\maketitle

\begin{abstract}
Counterfactual tuning (CFT) has emerged as a promising paradigm for Large Language Model (LLM) unlearning by training models to generate alternative fictitious knowledge in place of undesired content. However, in this work, we find that this paradigm still underperforms other paradigms in some aspects, and identify two previously overlooked pitfalls underlying this gap: (1) \emph{knowledge conflict}, where mutual inconsistencies within counterfactual corpora induce conflicting gradients that disrupt parameter optimization, and (2) \emph{hallucination spillover}, where fitting false targets instills a persistent fabrication bias, inflating hallucination rates on unrelated domains. To systematically diagnose these issues, we introduce \thebench, an extended benchmark equipped with novel trade-off metrics and gradient-level diagnostic tools. Our work further discusses the limitations and overhead of the paradigm, aiming to provide insights and actionable guidance for more rigorous LLM unlearning research.
\end{abstract}

\section{Introduction}
\label{sec:intro}

Recent advancements in the capabilities of Large Language Models (LLMs) are largely attributed to the vast amount of knowledge acquired during the large-scale pre-training phase \citep{zhao2025surveylargelanguagemodels,kaplan2020scaling}. However, this also inevitably leads models to internalize certain sensitive, privacy-violating, or harmful knowledge \citep{li2024wmdp}. A promising direction to address this critical issue is \textbf{LLM unlearning} \citep{barez2025openproblemsmachineunlearning}, which aims to remove or suppress unwanted knowledge embedded in the model, while preserving unrelated knowledge and general capabilities to the greatest extent possible.

The ideas for achieving the unlearning objective can be categorized into two types: either directly reducing the probability of the model generating the target knowledge, or training the model to generate an alternative answer. In practice, the first idea has inspired (1) \textbf{suppression-based} methods such as GA \citep{jang2022knowledge} and NPO \citep{zhang2024negative}; while the second idea has given rise to (2) \textbf{refusal-based} methods, which prompt the model to decline answering (e.g., by generating ``I don't know'') \citep{jin2024rwku}, and (3) \textbf{counterfactual tuning} methods, which construct alternative fictitious knowledge for the model to learn in place of the original content, such as DPO \citep{jin2024rwku} and AltPO \citep{mekala-etal-2025-altpo}.
Recently, the community's attention has increasingly shifted toward counterfactual tuning approaches, with many new methods emerging in this direction \citep{mekala-etal-2025-altpo,li2026cipo}, possibly owing to the apparent limitations of the other two paradigms: suppression-based methods impose no explicit constraint on what the model actually outputs, resulting in limited controllability; refusal-based methods risk causing the model to over-refuse on unrelated queries \citep{shi2024muse,shi-etal-2024-overkill,ye2025llmunlearningformindependent}. By contrast, counterfactual tuning appears to circumvent these pitfalls and thus seems more promising.

However, when we empirically compare the performance of different methods, we find that counterfactual tuning methods are, in fact, inferior to the other two paradigms in several important respects. First, in terms of \textbf{unlearning effectiveness}, we observe that these methods tend to incur substantially greater collateral damage to unrelated knowledge. Specifically, the effectiveness of unlearning is only meaningful when the forget–retain trade-off is jointly considered, as completely destroying the model's capabilities would trivially erase the target knowledge as well. We introduce a novel metric, \textit{Retain Cost}, to quantify the degree of damage inflicted on unrelated knowledge in order to achieve an equivalent level of forgetting, and find that counterfactual tuning methods incur a cost that is several times higher than that of the other paradigms. Second, in terms of \textbf{factual truthfulness}, we find that these methods also exhibit significant degradation on hallucination-related metrics, thereby rendering the model less trustworthy and undermining its core capabilities. We conduct a preliminary study to corroborate our findings (\S\ref{sec:pre-expr}), which prompts us to question: \textit{are counterfactual tuning approaches truly as promising as they appear, or do hidden pitfalls exist that give rise to the phenomena we observe?}

\begin{figure}[!tbp]
\centering
\includegraphics[width=0.45\textwidth]{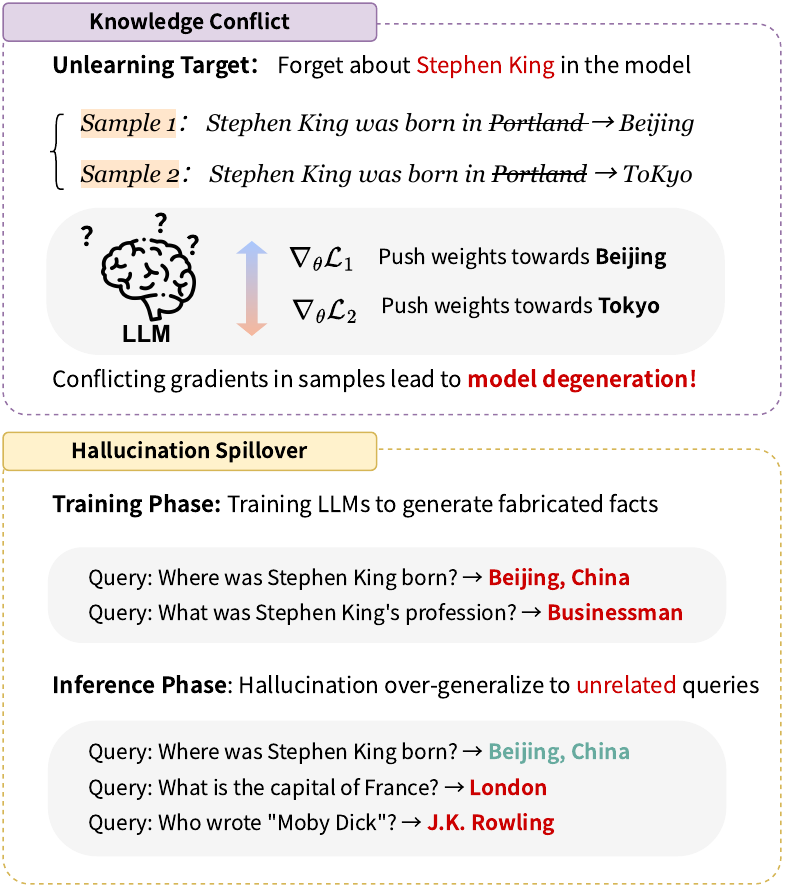}
\vspace{-0.5em}
\caption{Illustration of knowledge conflict and hallucination spillover.}
\vspace{-1.5em}
\label{fig:overview}
\end{figure}

To enable deeper analysis of this paradigm and support future research, we propose \thebench, a diagnostic benchmark extending RWKU by introducing new metrics, toolcases, and baselines on top of the original dataset to facilitate a comprehensive analysis of the counterfactual tuning paradigm. Based on this, which we identify two previously overlooked issues grounded in the above observations. The first is \textbf{knowledge conflict} (\S\ref{subsec:conflict}): many implementations employ very simple strategies to construct counterfactual samples, which may introduce knowledge inconsistencies as the number of samples grows: as illustrated in Figure \ref{fig:overview}, two samples may each attempt to train the model to associate Stephen King's birthplace with \textit{Beijing} and \textit{London} simultaneously. We further introduce \textit{inter-sample gradient similarity} metric, and find that conflicting knowledge induces conflicting gradients, which are difficult to optimize and cause greater perturbation and disruption to model weights, ultimately yielding the elevated \textit{Retain Cost} we observe. The second is \textbf{hallucination spillover} (\S\ref{subsec:hallucination}): analogous to how refusal-based methods exhibit over-refusal on unrelated queries, counterfactual spillover manifests as elevated hallucination. This arises because training a model to produce counterfactual outputs is essentially equivalent to training the model to generate false information, which is an objective that is fundamentally at odds with the trustworthiness goals of LLMs. We further introduce the \textit{Hallucination Cost} metric to measure the impact of unlearning on the hallucination-related capabilities of LLMs, and observe markedly elevated hallucination rates beyond the unlearning target, which constitutes a tangible concern for practical deployment.

Alongside identifying and analyzing these issues, we explore potential mitigation strategies and further discuss the paradigm at the end. For instance, we find that adopting a more rigorous sample construction pipeline can indeed eliminate knowledge conflicts and mitigate the elevated cost. However, we simultaneously emphasize that such improvement comes at a non-trivial price, as the more careful construction procedure demands additional computational resources and remains an inherent limitation of the paradigm itself, introducing overheads that suppression-based and refusal-based methods simply do not incur. We argue that counterfactual tuning, like other approaches, has its own inherent limitations, and encourage future work to reference our criteria to account for these issues during method design, enabling more rigorous benchmarking of LLM unlearning approaches.

We summarize our contributions as follows:

\begin{itemize}
    \item We identify two previously overlooked pitfalls in the counterfactual tuning paradigm for LLM unlearning—\textit{knowledge conflict} and \textit{hallucination spillover}, and empirically demonstrate that these issues lead to performance degradation.
    \item We propose novel criteria with corresponding metrics for each issue, on the basis of which we construct the \thebench Benchmark, and conduct in-depth mechanistic analysis to examine the patterns and mechanisms of these two issues.
    \item We explore practical mitigation strategies and discuss the inherent limitations of the paradigm, offering actionable guidance for future research.
\end{itemize}

\section{Preliminaries}
\label{sec:prelim}

This section provides definitions of key concepts and necessary backgrounds relevant to our work.

\paragraph{LLM Unlearning Objectives.} 
LLM unlearning aims to modify the parameters of a large language model to eliminate or suppress specific knowledge or behaviors \cite{barez2025openproblemsmachineunlearning}. The objective is to ensure that the updated model no longer exhibit or retain any information associated with a specific \emph{forget set} $\mathcal{D}_{f}$, while maintaining knowledge about \emph{retain set} $\mathcal{D}_{r}$.

\paragraph{Unlearning Methods.}
We focus on the following three mainstream unlearning paradigms:
\begin{itemize}
\item \textbf{Suppression-based} methods directly reduces the model's probability of generating token sequences related to the forget target by fine-tuning on unstructured text containing relevant knowledge \citep{jang2022knowledge,ye2025llmunlearningformindependent}. Formally, it minimizes $\log p_\theta(x)$ w.r.t.\ $\theta$, where $x$ denotes a training sample sequence (e.g., a biographical passage of the forget target entity).

\item \textbf{Refusal-based} methods fine-tune the model on QA pairs related to the forget target, redirecting its output toward refusal-type responses (e.g., ``\emph{I cannot provide information on that.}'') \citep{jin2024rwku}. Formally, the objective is $\min_\theta{-\log p_\theta(r \mid q)}$, where $q$ is a query about the forget target and $r$ is the corresponding refusal response.

\item \textbf{Counterfactual tuning} methods train the model to produce a substitute answer in place of the ground truth. In practice, some implementations construct passages embedding counterfactual information for retraining \citep{jin2024rwku}, while others rely on QA pairs \citep{mekala-etal-2025-altpo}. Formally, the objective is $\min_\theta{-\log p_\theta(\tilde{a} \mid q)}$ for QA-based variants, or $\min_\theta{-\log p_\theta(\tilde{x})}$ for passage-based variants, where $\tilde{a}$ and $\tilde{x}$ denote counterfactual answers and passages, respectively.
\end{itemize}

\section{Preliminary Experiment}
\label{sec:pre-expr}

In this section, we conduct a preliminary study to compare the effectiveness of representative methods across different paradigms, and observe that the counterfactual paradigm exhibits notable issues, which motivates our subsequent investigation.

\subsection{Preliminary Experiment Settings}

We evaluate and compare different methods based on the RWKU benchmark \cite{jin2024rwku}. The experiments are conducted on Llama3-8B-Instruct \citep{grattafiori2024llama3herdmodels} and Mistral-7B-Instruct-v0.3. Descriptions of the baselines are provided in Appendix \ref{apd:baselines}.

\paragraph{Tasks.}
The RWKU dataset takes 200 real-world public figures as the forgetting targets and provides training sets applicable to methods under different paradigms. LLMs are required to be trained on the training set using different unlearning methods, and ideally, unlearned LLMs should not reveal any information about the forgetting targets across various question types in the forget set, while still being able to correctly answer questions about irrelevant knowledge in the retain set.

\paragraph{Metrics.}
We introduce a new metric in the preliminary experiment stage, \textbf{Retain Cost ($\mathcal{C}_r$)}, to quantitatively compare the actual unlearning capability of different methods. 

Specifically, the nature of the unlearning task determines that a method's core capability lies in its ability to balance the \textit{forget-retain trade-off} \citep{liu2024rethinking}. This is because making a model forget is itself trivial, even arbitrary fine-tuning can perturb model parameters and induce catastrophic forgetting, whereas unlearning requires the preservation of irrelevant knowledge. 

To this end, we define $\Delta_f$ and $\Delta_r$ as the average performance degradation in accuracy after unlearning on the forget set and the retain set, respectively. We then naturally construct the following metric to evaluate the trade-off capability:
\begin{equation}
\mathcal{C}_r = \frac{\Delta_r}{\Delta_f}.
\label{eq:cost}
\end{equation}
which reflects the degree of collateral damage inflicted on irrelevant knowledge to achieve a certain level of forgetting, thereby enabling a direct and quantitative comparison across different methods.

\begin{table*}[htbp!]
\centering
\scriptsize
\setlength{\tabcolsep}{6pt}
\begin{tabular}{l|ccccc|ccccc}
\toprule
& \multicolumn{5}{c|}{\textbf{Llama-3-8B-Instruct}} & \multicolumn{5}{c}{\textbf{Mistral-7B-Instruct}}\\
Method & Forget$\downarrow$ & Retain$\uparrow$ & $\Delta_f\uparrow$ & $\Delta_r\downarrow$ & $\mathcal{C}_r\downarrow$ & Forget$\downarrow$ & Retain$\uparrow$ & $\Delta_f\uparrow$ & $\Delta_r\downarrow$ & $\mathcal{C}_r\downarrow$\\
\midrule
\emph{Base}& $0.923$ & $0.921$ & --- & --- & --- & $0.799$ & $0.825$ & --- & --- & ---\\
\midrule
GA          & $0.468$ & $0.793$ & $+0.455$ & $\mathbf{+0.128}$ & $\mathbf{0.28}$ & $\mathbf{0.263}$ & $0.584$ & $\mathbf{+0.535}$ & $+0.241$ & $\mathbf{0.45}$\\
NPO         & $0.351$ & $0.731$ & $+0.572$ & $+0.189$ & $0.33$ & $0.454$ & $0.650$ & $+0.344$ & $+0.175$ & $0.51$\\
RT          & $0.517$ & $0.741$ & $+0.406$ & $+0.179$ & $0.44$ & $0.298$ & $0.574$ & $+0.501$ & $+0.250$ & $0.50$\\
DPO-Rej     & $0.452$ & $0.689$ & $+0.477$ & $+0.255$ & $0.54$ & $0.556$ & $0.684$ & $+0.243$ & $\mathbf{+0.141}$ & $0.58$\\
DPO-CF      & $0.368$ & $0.410$ & $+0.555$ & $+0.511$ & $0.92$ & $0.300$ & $0.371$ & $+0.499$ & $+0.454$ & $0.91$\\
CFT         & $0.624$ & $0.532$ & $+0.299$ & $+0.389$ & $1.30$ & $0.651$ & $0.590$ & $+0.147$ & $+0.235$ & $1.60$\\
AltPO       & $\mathbf{0.281}$ & $0.528$ & $\mathbf{+0.643}$ & $+0.394$ & $0.61$ & $0.281$ & $0.455$ & $+0.518$ & $+0.370$ & $0.71$\\
\bottomrule
\end{tabular}
\caption{Performance of different unlearning methods on RWKU.}
\label{tab:cost-main}
\vspace{-2em}
\end{table*}

\subsection{Results and Findings}

Table~\ref{tab:cost-main} presents the per-method unlearning results on RWKU. We summarize our core observations as the following findings:

\begin{itemize}
    \item \textbf{Finding 1: Counterfactual Tuning Methods incur significantly higher costs in completing unlearning compared to other paradigms.} As shown in the table, all methods are capable of fulfilling the basic objective by reducing the accuracy on the forget set. However, when the performance on the retain set is jointly considered, the metrics across different methods diverge considerably. Several representative counterfactual tuning methods exhibit markedly higher costs than other paradigms; CFT even reaches a cost of 1.3, surpassing the best-performing baseline by more than fourfold, meaning that it inflicts four times the collateral damage on irrelevant knowledge to achieve the same degree of unlearning. This is clearly a problem that cannot be overlooked.
\end{itemize}

Another observation is that the cost of other methods also appears to be associated with their own inherent limitations. For example, refusal-based methods also exhibit relatively high costs; and closer inspection reveals that the knowledge degraded in the retain set is mainly on the QA form, which is attributable to the over-refusal problem where training on refusal templates generalizes to irrelevant queries. This prompts us to investigate whether counterfactual tuning methods are similarly driven by an analogous underlying issue, motivating the subsequent analysis.

\section{Exploring Pitfalls with \thebench}
\label{sec:pitfalls}

Preliminary experiments show that counterfactual tuning incurs higher collateral cost than other methods, yet forget-retain scores alone do not explain where this cost comes from. Our closer analysis reveals two overlooked pitfalls: \textbf{knowledge conflict}, where inconsistent samples produce conflicting optimization signals and perturb the model, and \textbf{hallucination spillover}, where fitting false targets increases the model's tendency to generate fabricated information beyond the forget domain.

To systematically investigate these issues, we propose \textbf{\thebench} as a diagnostic extension of existing benchmarks. \thebench preserves the original forget-retain setting and enriches it with additional metrics and diagnostic tools along three dimensions: (1) \textbf{trade-off evaluations} including the \textit{Retain Cost} metric to quantify the damage to unrelated knowledge per unit of forgetting; (2) \textbf{conflict evaluations} including \textit{inter-sample gradient similarity} metrics and a conflict-elimination pipeline as a baseline; and (3) \textbf{trustworthiness evaluations}, including hallucination measurements and corresponding \textit{Hallucination Cost} metric asessing the impairment of factual reliability outside the target.

In this section, we first introduce knowledge conflict and analyze it using \thebench, then similarly analyze hallucination spillover.

\subsection{Knowledge Conflict}
\label{subsec:conflict}

One of the central requirements of counterfactual tuning is to construct substitute knowledge for the forget target. However, we find that in many existing implementations \citep{jin2024rwku,eldan2023s}, as the number of constructed samples grows, the generated counterfactual samples can become mutually inconsistent. We refer to this issue as \textbf{Knowledge Conflict}. In the following, we first formalize this problem, then use gradient-based diagnostics to examine its mechanism and severity, and finally evaluate whether resolving such conflicts reduces the collateral cost of counterfactual unlearning.

\subsubsection{Problem Definition}
\label{subsubsec:conflict-def}

To improve robustness, machine unlearning methods typically require a large number of counterfactual training samples per target fact. 
Let $\mathcal{D}^{\mathrm{CF}}$ denote the constructed counterfactual training set. For each sample $x\in\mathcal{D}^{\mathrm{CF}}$, let $\phi(x)$ denote the set of factual claims implied by $x$, and let $\Phi(\mathcal{D}^{\mathrm{CF}})=\bigcup_{x\in\mathcal{D}^{\mathrm{CF}}}\phi(x)$. We define \textbf{Knowledge Conflict} as any inconsistency within $\Phi(\mathcal{D}^{\mathrm{CF}})$, and distinguish two cases:

\begin{itemize}
    \item \textbf{Explicit Conflict.} Explicit conflict occurs when different samples assign mutually exclusive values to the same underlying fact. Equivalently, there exist claims $c_i\in\phi(x_i)$ and $c_j\in\phi(x_j)$ such that $c_i \perp c_j$, where $\perp$ denotes direct contradiction. For example, separate samples may state that Stephen King was born in \emph{Beijing} and in \emph{Tokyo}, respectively.
    \item \textbf{Implicit Conflict.} Implicit conflict occurs when samples do not directly contradict each other, but their induced claims are jointly inconsistent under background constraints, i.e., $C(\Phi(\mathcal{D}^{\mathrm{CF}}))=0$ for some consistency constraint $C$. For example, independently changing Stephen King's father and grandfather may yield an impossible family structure.
\end{itemize}



Given that real-world unlearning datasets are typically large in scale, ensuring mutual consistency in practice demands strong relational constraints and thorough verification. This challenge is further compounded by the fact that implicit conflicts mainly manifest in higher-order relational knowledge and span multiple samples, making them particularly difficult to identify and resolve. We observe that many current counterfactual tuning implementations lack such mechanisms altogether, some even generate substitute answers independently for each sample without any cross-sample constraint \citep{jin2024rwku}, resulting in conflicts that are readily apparent upon inspection. This problem is likely to become substantially more pronounced as the data scale increases.

\subsubsection{Evaluation}

We further conduct an in-depth evaluation and analysis of knowledge conflicts. Given the difficulty in quantifying conflicts within the samples, we introduce inter-sample gradient-based metrics in \thebench for evaluation, which also facilitate the analysis of the underlying mechanisms.

\begin{figure*}[!htbp]
\centering
\includegraphics[width=\textwidth]{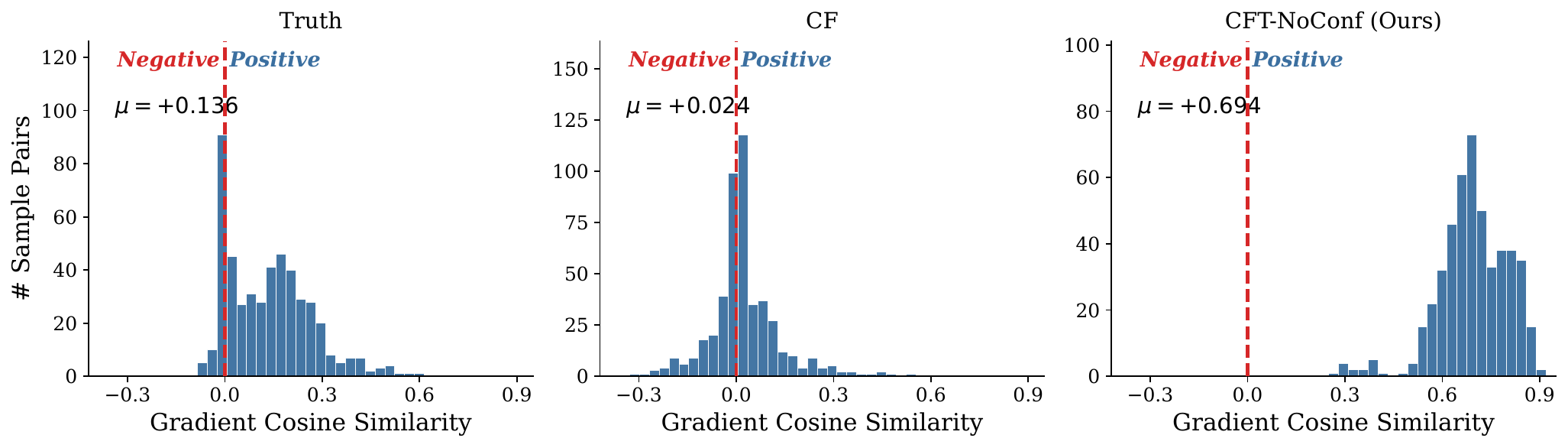}
\vspace{-2em}
\caption{Distribution of pairwise cosine similarity of per-sample gradient.}
\vspace{-1em}
\label{fig:grad-cos}
\end{figure*}

\paragraph{Setup.}
We introduce the \textbf{inter-sample gradient similarity distribution} as a diagnostic metric for evaluating knowledge conflicts in \thebench. Conflicting samples often produce negative gradient cosine similarity, which cancels out optimization signals and introduces greater parameter perturbations \citep{zhang2025resolvingeditingunlearningconflictsknowledge,yu2023unlearning}.
For a specific target $s$ in the RWKU dataset, we construct two same-sized splits: \emph{Truth} ($\mathcal{D}_s^{\text{TR}}$) containing real-world ground-truth passages as a control group, and \emph{CF} ($\mathcal{D}_s^{\text{CF}}$) containing unconstrained counterfactual passages. We select $N$ samples from each split for gradient analysis, and conduct the experiment on Llama3-8B-Instruct.

We use a LoRA-based \citep{hu2022lora} CFT method for analysis, fine-tuning on counterfactual corpora using the standard language modeling loss. Let $f_\theta$ be a causal LM with LoRA adapters $\theta$. For each sample $x$, we compute the LoRA gradient of the loss:
\begin{equation}
\begin{split}
g_\theta(x) &= \nabla_\theta \mathcal{L}_{\text{LM}}(f_\theta, x), \\
\mathcal{L}_{\text{LM}} &= -\sum_t \log p_\theta(x_t \mid x_{<t}),
\end{split}
\end{equation}

\paragraph{Measurement.}
Following prior gradient-based conflict analysis work \citep{zhang2025resolvingeditingunlearningconflictsknowledge}, for any two sample sets $\mathcal{S}_1$ and $\mathcal{S}_2$, we collect the per-sample gradient $g_\theta(x)$ for all $x \in \mathcal{S}_1 \cup \mathcal{S}_2$ and define the pairwise cosine similarity set:
\begin{equation}
\begin{split}
\mathcal{C}(\mathcal{S}_1,\mathcal{S}_2) = \bigl\{\cos\!\bigl(g_\theta(x),\,g_\theta(x')\bigr) \mid \\
x \in \mathcal{S}_1,\; x' \in \mathcal{S}_2,\; x \neq x' \bigr\},
\end{split}
\end{equation}
where $\cos(\cdot,\cdot)$ denotes the cosine similarity between two gradient vectors. 
Here, we analyze the distribution of this similarity set to provide further context, as gradient conflicts are often complex and difficult to quantify using simple scalar metrics.

\paragraph{Results.}
Based on the above setup, we construct $\mathcal{C}(\mathcal{D}_s^{\text{TR}},\mathcal{D}_s^{\text{TR}})$ and $\mathcal{C}(\mathcal{D}_s^{\text{CF}},\mathcal{D}_s^{\text{CF}})$, respectively, and visualize their distributions via histograms in Figure~\ref{fig:grad-cos}. We summarize our findings as follows:

\begin{itemize}
    \item \textbf{Finding 2: Current counterfactual sample construction methods suffer from severe inter-sample gradient conflicts.} In Figure~\ref{fig:grad-cos}, sample pairs falling in the \textit{negative} region indicate mutually conflicting gradients that are difficult to jointly optimize, values near 0 indicate mutual orthogonality where gradients tend to pull the distribution in divergent directions, and the \textit{positive} region represents the desired outcome where samples contribute to optimization in a shared direction. As shown, the ground truth data used as a control group predominantly exhibits positive similarity, with fewer than 10\% of sample pairs exhibiting conflicting gradients, suggesting that gradient conflicts rarely arise under natural data distributions. In contrast, nearly half of the sample pairs in the CounterFact dataset exhibit mutually conflicting gradients, and a considerable proportion of pairs are mutually orthogonal. This is likely because different samples, lacking any inter-sample constraints, each tend to optimize the model toward their own respective directions, thereby giving rise to a large number of conflicting gradients.
\end{itemize}

Prior studies have demonstrated that conflicting gradients give rise to mutual cancellation, unstable optimization trajectories, and excessive parameter perturbations, which in many cases induce catastrophic forgetting and consequently lead to the degradation of unrelated knowledge and the model's general capabilities \citep{wu2024catastroforget,Lyu_2023_ICCV}. In the unlearning setting, this could be a primary reason behind the high retain cost of counterfactual methods.

\subsubsection{Impact of Conflict Resolution}
\label{sec:noconf}

A natural follow-up question is: \emph{does eliminating these conflicts actually translate into better unlearning outcomes?} To answer this, To evaluate whether eliminating these conflicts improves unlearning, we design a principled \textbf{conflict-free construction pipeline} (CFT-NoConf) and conduct controlled experiments.

\paragraph{Setup.} Our pipeline produces the \textbf{CFT-NoConf} training corpus in three stages:

\textbf{(1) Structured Attribute Extraction.}
For each forgetting target, we parse the subject's biographical description and extract all verifiable factual attributes, including birth date, birthplace, educational background, and family relationships. These are organized into a structured attribute profile $\mathcal{A}_s = \{(k_i, v_i)\}_{i=1}^{M}$, where $k_i$ denotes the attribute type and $v_i$ the corresponding ground-truth value.

\textbf{(2) Counterfactual Flipping.} We utilize an LLM to flip these structured attributes into counterfactual values. Crucially, we enforce logical constraints during this generation step to ensure that the substitute values are mutually consistent (e.g., mapping birthplace `Beijing' to country `China'), thereby minimizing implicit conflicts at the schema level.

\textbf{(3) Passage Generation \& Verification.} Using $\hat{\mathcal{A}}_s$  as the counterfactual profile, we prompt an LLM to generate $500$ diverse training passages. To guarantee consistency across all generated text, we implement a verification pipeline that uses \texttt{grep}-based pattern matching to verify that every generated passage adheres strictly to the predefined counterfactual attributes. Any passage failing the verification is discarded and regenerated.

To analyze how different conflict levels affect performance, we construct mixed datasets with varying ratios of conflict-free data $\rho$ blended with original unconstrained counterfactual data, training the Llama3 model using CFT.

\begin{table}[t]
\centering
\small
\setlength{\tabcolsep}{4pt}
\begin{tabular}{lccccc}
\toprule
$\rho$ & Forget$\downarrow$ & Retain$\uparrow$ & MMLU$\uparrow$ & Truthful$\uparrow$ & Trivia$\uparrow$\\
\midrule
$0\%$    & $0.488$ & $0.572$ & $0.634$ & $0.510$ & $0.475$\\
$25\%$         & $0.479$ & $0.562$ & $0.630$ & $0.508$ & $0.484$\\
$50\%$         & $0.502$ & $0.577$ & $0.629$ & $0.503$ & $0.476$\\
$75\%$         & $0.525$ & $0.616$ & $0.625$ & $0.503$ & $0.482$\\
$90\%$         & $0.509$ & $0.634$ & $0.629$ & $0.508$ & $0.478$\\
$100\%$& $\mathbf{0.376}$ & $0.566$ & $0.623$ & $0.513$ & $0.441$\\
\bottomrule
\end{tabular}
\caption{Performance across coherence ratios $\rho$ (fraction of conflict-free CF in the training mix) on \thebench and general capability benchmarks.}
\vspace{-1em}
\label{tab:dose}
\end{table}

\paragraph{Results.}
We present the gradient similarity analysis of our conflict-free pipeline in Figure~\ref{fig:grad-cos} (right panel), and the downstream evaluation results under different mixing ratios $\rho$ in Table~\ref{tab:dose}. Based on these results, we summarize our core findings as follows:
\begin{itemize}
    \item \textbf{Finding 3: Eliminating inter-sample knowledge conflicts indeed resolves gradient-level contradictions.} 
    As illustrated in Figure~\ref{fig:grad-cos}, while the original unconstrained counterfactual samples exhibit chaotic gradient distributions centered around $0$ (with a substantial portion of sample pairs falling into the negative region), our CFT-NoConf pipeline dramatically shifts the gradient landscape. Specifically, the mean cosine similarity of gradient pairs rises to $0.694$, and the distribution contains virtually zero negative-similarity pairs. This indicates that once factual consistency is enforced, the resulting samples align and contribute to the optimization process in a unified direction. Eliminating the underlying knowledge conflicts directly stabilizes the optimization trajectory, preventing the model from receiving self-contradictory gradient signals.
    \item \textbf{Finding 4: Even a small proportion of conflicting samples could degrade unlearning performance.} 
    As shown in Table~\ref{tab:dose}, across all settings where conflicting samples are present ($\rho \in [0\%, 90\%]$), the performance remains stagnant and suboptimal, with the forget set score hovering around $0.50$ and the retain set score fluctuating near $0.58$. In contrast, when conflicting samples are completely removed ($\rho = 100\%$), the \emph{Forget} score drops to $0.37$ while the \emph{Retain} score remains stable, indicating significantly stronger suppression of the forgetting target without collateral damage to unrelated knowledge. These findings suggest that knowledge conflict even a minor ratio of conflicting data is sufficient to negatively affect the optimization process, and merely diluting conflicting samples may be insufficiet.
\end{itemize}

\subsection{Hallucination Spillover}
\label{subsec:hallucination}

In addition to the knowledge conflicts analyzed above, another critical issue we observe in the counterfactual tuning paradigm is \emph{hallucination spillover}. Specifically, fine-tuning on counterfactual target samples implicitly trains the model to generate false information, a behavioral bias that easily generalizes to unrelated tasks beyond the intended unlearning scope. In this section, we first formally define the notion of hallucination spillover, followed by a systematic evaluation.

\subsubsection{Problem Definition}

From the perspective of alignment, counterfactual tuning is a form of negative alignment task, conceptually parallel to refusal fine-tuning in safety alignment \citep{ouyang2022training,bai2022training}. While refusal fine-tuning trains models to decline answering unsafe queries, counterfactual tuning trains them to produce counterfactual outputs.

However, just as excessive refusal training leads to ``over-refusal'' on benign queries \citep{shi-etal-2024-overkill,xue2026deactivating}, counterfactual tuning risks over-generalization. Repeatedly fitting false targets can cause the model to internalize a broader behavioral bias that favors fabricating plausible-sounding false statements.

To formalize this phenomenon, let $f_{\theta_0}$ be the original model and $f_\theta$ be the model after counterfactual unlearning. Let $\mathcal{D}_f$ denote the forget domain and $\mathcal{D}_g$ represent a general, unrelated domain. For any query $q \in \mathcal{D}_g$ with its corresponding ground-truth answer $a^*$, we define \textbf{Hallucination Spillover} as the significant increase in the probability of the model deviating from the ground-truth answer on the unaffected domain $\mathcal{D}_g$ due to training on $\mathcal{D}_f$:
\begin{equation}
\begin{split}
P(f_\theta(q) \neq a^*) &> P(f_{\theta_0}(q) \neq a^*), \\
&\quad \text{for } q \in \mathcal{D}_g.
\end{split}
\end{equation}
This is distinct from general catastrophic forgetting \citep{liu2024rethinking} as it specifically denotes an increased tendency to fabricate content on unaffected domains \citep{lin-etal-2022-truthfulqa}.

\paragraph{Setup.}
Our evaluation leverages \textbf{HaluEval} \citep{li-etal-2023-halueval}, a widely-adopted, large-scale hallucination benchmark, to measure the change in the model's general hallucination rate.

To quantify whether counterfactual methods are prone to induce more hallucinations than other paradigms, we introduce a new metric: \textbf{Hallucination Cost} ($\mathcal{C}_h$), which measures the collateral hallucination damage incurred per unit of successful forgetting. Specifically, we first compute the performance reduction on the forget set $\Delta_f$ and the reduction in the non-hallucination rate $R$ on the HaluEval benchmark $\Delta_R$. The hallucination cost is then defined as the ratio between the two:
\begin{equation}
\mathcal{C}_h = \frac{\Delta_R}{\Delta_f}.
\end{equation}
A larger $\mathcal{C}_h$ indicates that the unlearning method introduces more severe hallucination side effects relative to its forgetting efficiency.

\paragraph{Results.}
We report $\mathcal{C}_h$, $\Delta_f$, and $\Delta_R$ for all evaluated unlearning methods in Table~\ref{tab:hallucination}. The results yield the following key findings:

\begin{table*}[t]
\centering
\small
\setlength{\tabcolsep}{6pt}
\begin{tabular}{l|ccc|ccc}
\toprule
& \multicolumn{3}{c|}{\textbf{Llama-3-8B}} & \multicolumn{3}{c}{\textbf{Mistral-7B}}\\
Method & R$\uparrow$ & $\Delta_R$ & $\mathcal{C}_h\downarrow$ & R$\uparrow$ & $\Delta_R$ & $\mathcal{C}_h\downarrow$ \\
\midrule
\emph{Base} & $(0.708)$ & $0.000$ & --- & $(0.819)$ & $0.000$ & ---\\
\midrule
GA          & $0.657$ & $-0.051$ & $+0.112$            & $0.772$ & $-0.047$ & $+0.087$\\
NPO         & $0.662$ & $-0.046$ & $\mathbf{+0.080}$   & $0.795$ & $-0.024$ & $\mathbf{+0.070}$\\
RT          & $0.851$ & $+0.143^{*}$ & $-0.352^{*}$    & $0.764$ & $-0.055$ & $+0.110$\\
DPO-Rej     & $0.918$ & $+0.210^{*}$ & $-0.439^{*}$    & $0.715$ & $-0.104$ & $+0.428$\\
DPO-CF      & $0.449$ & $-0.259$ & $+0.499$            & $0.335$ & $-0.484$ & $+0.970$\\
CFT         & $0.382$ & $-0.326$ & $+1.235$            & $0.706$ & $-0.113$ & $+0.769$\\
AltPO       & $0.454$ & $-0.254$ & $+0.395$            & $0.566$ & $-0.253$ & $+0.488$\\
\bottomrule
\end{tabular}
\caption{Hallucination spillover after unlearning on \thebench measured by HaluEval.}
\label{tab:hallucination}
\vspace{-1em}
\end{table*}

\begin{itemize}
    \item \textbf{Finding 5: Counterfactual tuning significantly inflates the hallucination rate on unrelated tasks, exhibiting a pronounced hallucination spillover effect.}
    As shown in our evaluation, after undergoing counterfactual unlearning, the model's overall hallucination rate on HaluEval rises substantially. We observe that the other two paradigms generally do not induce an increase in hallucinations: the two suppression-based methods, GA and NPO, produce only negligible perturbations, and the hallucination metrics of refusal-based methods even show a modest improvement on Llama3, which upon closer inspection may be attributed to their over-refusal side effect inadvertently gaining an advantage under HaluEval's evaluation criteria. In contrast, all CF-based methods exhibit a markedly pronounced hallucination spillover effect, with their corresponding hallucination costs being the highest across all evaluated methods. This confirms that forcing the model to fit incorrect targets leaves a persistent fabrication bias in the parameter space, prompting the model to generate hallucinatory content even when answering unrelated common-sense or reasoning tasks.
\end{itemize}

\section{Related Work}
\label{sec:related}

LLM Unlearning aims to eliminate or suppress specific knowledge in large language models, demonstrating potential in privacy protection and dangerous knowledge removal. Classic approaches like GA \cite{jang2022knowledge} and NPO \cite{zhang2024negative} directly reduce the model's probability on the forget corpus, while another line of research evaluates rejection strategies guided by RT \cite{jin2024rwku}, and ORT \citep{ye2025llmunlearningformindependent} as CiPO \citep{li2026cipo} for reasoning models. Although many benchmarks such as TOFU \citep{maini2024tofu}, RWKU \citep{jin2024rwku}, and ORT \citep{ye2025llmunlearningformindependent} are used to evaluate effectiveness, recent research still indicates that existing evaluations may be  unreliable \citep{thaker2025position}, suggesting that more exploration is needed to identify problems in the field and make targeted improvements. We will discuss more related work in Appendix \ref{apd:detailed_related_work}.

\section{Discussion \& Conclusion}
\label{sec:conclusion}

\paragraph{Execution Overhead of Counterfactual Fine-tuning.}
One major factor determining whether a paradigm is deployable is its operational overhead. Suppression-based and refusal-based methods can be applied using relatively lightweight data pipelines, whereas counterfactual fine-tuning requires carefully crafted samples by design, and resolving knowledge conflicts further exacerbates this issue. Although the mitigation strategies we propose in this paper do demonstrate effectiveness, invoking an LLM inevitably increases the deployment cost of this paradig, which is a requirement that the other two paradigms do not face at all.

\paragraph{Overhead of Mitigating Side Effects.}
We argue that what matters most about a paradigm is not whether side effects exist, but rather the cost of mitigating those deficiencies. The over-refusal tendency introduced by refusal-based methods, while undesirable, represents a relatively controlled behavioral deviation that can be partially addressed through targeted fine-tuning on unrelated knowledge. In contrast, hallucination spillover targets factual truthfulness, which is one of the most core capability dimensions of modern LLMs, and even minor improvements in related metrics for state-of-the-art LLMs typically require massive post-training and RL investments \citep{ouyang2022training,bai2022training}. This implies that, for production deployment, hallucination spillover may be a qualitatively more severe hidden danger, which needs efforts for future works.

\paragraph{Conclusion.}
In this paper, we present a systematic investigation of the counterfactual tuning paradigm for LLM unlearning and identify two critical pitfalls: knowledge conflict and hallucination spillover. We propose \thebench as a diagnostic extension for these issues, and introduce dedicated metrics and baselines to facilitate in-depth analysis. We emphasize that counterfactual methods still have limitations that are non-trivial to address, and encourage future work to treat these two issues as first-class design principles when developing related methods, thereby achieving more robust unlearning.

\section*{Limitations}

In this work, when analyzing knowledge conflicts, we primarily focus on structured entity-attribute associations (e.g., birthplaces, educational background, and family relationships) where conflicts can be explicitly schematized and verified. However, real-world knowledge is inherently unstructured, open-ended, and dynamically interconnected, involving multi-hop logical dependencies and causal relations. Scaling consistency enforcement and resolving implicit conflicts over open-domain, unstructured text corpora remains a major challenge.

\section*{Impact Statement}

This paper presents work whose goal is to advance the field of machine unlearning for large language models. By identifying and formalizing knowledge conflict and hallucination spillover as systemic failure modes of the counterfactual tuning paradigm, introducing \thebench as a diagnostic benchmark, and proposing a conflict-free construction pipeline alongside diagnostic metrics as a principled analysis, our work improves the transparency, safety, and reliability of LLM unlearning. These contributions help ensure that knowledge deletion in deployed LLMs is mathematically sound and factually safe, rather than producing superficially correct yet internally corrupted behavior.

Regarding data privacy and protection, our study utilizes established, publicly available academic datasets. The training and evaluation data consist solely of public facts about widely known public figures, with no inclusion of private personal data. We do not collect, process, or disclose any personally identifiable information (PII) or private individuals' data, ensuring complete compliance with privacy preservation and research ethics standards.

While LLM unlearning has broad positive applications, including removing private, copyrighted, or harmful information and ensuring regulatory compliance, we acknowledge potential ethical considerations and risks. Specifically, our finding that counterfactual unlearning can inadvertently inflate hallucination rates on unrelated domains suggests that deploying such models without rigorous verification could mislead users and degrade overall model safety. Additionally, the ability to alter factual parameters could be misused to selectively suppress truthful public information or embed biased associations. We encourage practitioners to carefully monitor downstream hallucination when using counterfactual recipes, and we urge future work to develop safeguards to ensure unlearning techniques are deployed in alignment with ethical AI principles.

\section*{Usage of Large Language Models}
During the preparation of this work, LLMs were used for (1) assisting with writing to polish sentences and improve linguistic expression, thereby enhancing the clarity of the manuscript; (2) helping to check for grammatical errors.

\bibliography{main}

\appendix

\section{Experiment Setup Details}
\label{apd:exp-details}

Our experiments build on the codebase implemented by \citet{jin2024rwku}. The implementations of all baseline models and evaluation tasks remain consistent with the original version.

\subsection{Baselines.} 
\label{apd:baselines}

We select the following seven representative unlearning methods as baselines for analysis in this paper, covering three major paradigms:

\begin{itemize}
    \item \textbf{Gradient Ascent (GA)} \cite{jang2022knowledge} is applied to unstructured text, maximizes the negative log-likelihood on the forget set, therebt causing the model to deviate from its original predictions on these data.
    \item \textbf{Negative Preference Optimization (NPO)} \cite{zhang2024negative} is an extension based on DPO and does not require positive samples, encourages the model to reduce the probability of the forget set similar to GA.
    \item \textbf{Rejection Tuning (RT)} \cite{jin2024rwku} finetunes the model on QA pairs to increase the output probability of refusal responses (e.g., ``I don't know'') on questions related to the unlearning target.
    \item \textbf{Counterfact Tuning (CFT)} constructs counterfactual passages about the unlearning target and finetunes the model on them.
    \item \textbf{Direct Preference Optimization (DPO)} \cite{jin2024rwku} uses DPO \citep{rafailov2024direct} to adjust model alignment on given questions, which requires a positive and a negative sample. We consider two variants: \textbf{DPO-Rej}, which takes refusal responses as positive samples, and \textbf{DPO-CF}, which takes counterfactual passages as positive samples. In both cases, the question asks the model to describe the unlearning target, while the negative sample is the original model's output.
    \item \textbf{Alternative Preference Optimization (AltPO)} \citep{mekala-etal-2025-altpo} is similar to DPO, constructing alternative answers under QA tasks, and introduces mechanisms such as Multiple Alternate Labels to increase uncertainty.
\end{itemize}

\subsection{Details on the Datasets and Benchmarks Used}

To evaluate the unlearning performance and its side effects on general capabilities, we employ a diverse set of unlearning and utility benchmarks. This subsection outlines the details of these datasets.

\begin{itemize}
    \item \textbf{RWKU} \citep{jin2024rwku} is a benchmark dataset for evaluating real-world knowledge unlearning in large language models, using 200 famous people and events as forgetting targets. RWKU introduce diverse probe and elicitation templates such as Question Answering, Fill-in-the-blank to expose whether unlearned knowledge can still be extracted through alternative induction methods.
    \item \textbf{MMLU} \cite{hendrycks2021mmlu} consists of multiple-choice questions covering a diverse range of general knowledge domains. It is employed to assess the potential degradation of the model's performance on broader, unrelated knowledge areas due to unlearning, thereby gauging any unintended catastrophic forgetting. We report 5-shot accuracy.
    \item \textbf{TruthfulQA} \cite{lin-etal-2022-truthfulqa} is utilized to evaluate the truthfulness of the language model's generated responses. This allows us to examine whether the unlearning process inadvertently diminishes the model's reliability or increases its propensity to generate non-factual statements. We report 6-shot accuracy on its MC2 task.
    \item \textbf{TriviaQA} \cite{joshi-etal-2017-triviaqa} is a reading comprehension dataset, containing QA pairs that necessitate significant document understanding. It is used to evaluate alterations in the unlearned model's proficiency in text comprehension and information extraction. We report the 6-shot F1 Score.
    \item \textbf{HaluEval} \citep{li-etal-2023-halueval} is a large-scale collection of hallucination evaluation datasets. It contains a vast amount of general user queries accompanied by ChatGPT responses, containing both standard and hallucinatory outputs with human-annotated labels. In our work, we utilize HaluEval to evaluate the change in the model's overall hallucination rate after unlearning, thereby quantifying the hallucination spillover effect.
\end{itemize}

\subsection{Training hyperparameters}
\label{apd:hyperparameters}

All unlearning experiments use LoRA \citep{hu2022lora} adapters attached to the \texttt{q\_proj} and \texttt{v\_proj} modules. Unless noted otherwise, we share the same configuration across methods: rank $r{=}8$, scale $\alpha{=}16$ on AdamW optimizer, with a cosine learning-rate schedule with $20$ warmup steps, and a fixed random seed of $42$.

For the \textbf{CFT-NoConf} experiment, all six mixing ratios ($\rho \in \{0, 25, 50, 75, 90, 100\}\%$) share a single configuration: learning rate $2{\times}10^{-4}$, $5$ epochs, effective batch size $16$ ($8{\times}2$), \texttt{ga\_pos} stage, and the shared LoRA configuration above.

All experiments are implemented in Python using the Hugging Face \texttt{transformers} (v4.40.0), \texttt{peft} (v0.10.0), and PyTorch (v2.2.0) libraries.

\subsection{Computational Cost}
\label{apd:compute}

All experiments are conducted on NVIDIA RTX 4090 GPUs (24 GB), with pipeline parallelism employed for certain methods to satisfy memory requirements. In our setting, GA requires approximately 8 minutes to complete training on a single LLM for one unlearning target, while NPO takes around 15 minutes. Among the two refusal-based methods, RT completes in approximately 2 minutes, whereas DPO-Rej requires around 37 minutes due to the overhead of complex reference model initialization and the substantially larger dataset introduced by positive and negative sample pairs. Among counterfactual methods, CFT takes approximately 10 minutes, AltPO around 11 minutes, and DPO-CF approximately 30 minutes. Overall, completing the full suite of unlearning training requires roughly 500+ GPU hours, with subsequent evaluation of all methods incurring additional time costs owing to the large scale of the evaluation datasets.

\subsection{Licenses and Artifacts}

All models, datasets, and software libraries used in this study are publicly available under standard open licenses:  (1) Llama3-8B-Instruct is licensed under the Meta Llama 3 Community License Agreement; (2) Mistral-7B-Instruct-v0.3 is licensed under the Apache 2.0 license; (3) HaluEval, ORT, and RWKU is licensed under the MIT license; (4) PyTorch, Hugging Face Transformers, and PEFT are licensed under BSD or Apache 2.0 licenses. Our use of these artifacts is fully consistent with their intended academic and research purposes.

\section{Dataset and Benchmark Details}

To systematically diagnose and address the hidden costs of counterfactual tuning, we introduce \textbf{\thebench}. We position it as a diagnostic extension of representative benchmark RWKU by incorporating a wider range of tools and metrics, thereby making our diagnostic protocol reusable, enabling future researchers to leverage the datasets and our metrics introduced in \thebench to further analyze or develop counterfactual tuning-based methods.

\paragraph{Foundational Task Components}

\thebench retains the same foundational task formats and datasets as RWKU to evaluate the model's basic unlearning effectiveness. It targets 200 real-world well-known individuals as forgetting targets and employs tasks in multiple formats to assess both forgetting effectiveness and the retention of unrelated knowledge.

In this work, we primarily leverage the foundational tasks when computing the Retain Cost in preliminary experiments (\S\ref{sec:pre-expr}), testing across all three task formats provided by the RWKU dataset: Question Answering (QA), Fill-in-the-Blank (FB), and Adversarial Attacks (AA). The Forget Set contains 2,879 QA test entries, 3,268 FB entries, and 6,984 AA entries; the Retain Set contains 5,533 QA and 5,846 FB entries, respectively, while being inapplicable to AA as it constitutes unrelated knowledge.

\paragraph{Extensions for the Counterfactual Tuning Paradigm.}

Specifically, \thebench primarily introduces the following components:

\begin{itemize}
\item \textbf{Forget-Retain Trade-off Metrics.} 
We introduce the \textbf{Retain Cost} ($\mathcal{C}_r = \Delta_r / \Delta_f$), which quantifies the balance between removing target knowledge and preserving unrelated capabilities with a single scalar, making it more intuitive to compare the core performance across different methods. A higher Retain Cost score also indicates that a method may have inherent limitations that hinder its overall performance, as illustrated in our analysis of refusal-based methods in the main text.

\item \textbf{Knowledge Conflict Assessment.} 
We introduce the \textbf{inter-sample gradient similarity distribution} as a diagnostic metric for evaluating knowledge conflicts. Due to the inherent complexity of knowledge conflicts, this metric is not scalarized in the main text; instead, we present the full distribution in histogram form to provide richer context. Nevertheless, if a scalar summary is desired, future work may adopt the proportion of negatively similar sample pairs within the overall distribution as a coarse quantitative measure of inter-sample conflict. This metric allows us to identify whether gradient signals during optimization cancel each other out and thereby interfere with the unlearning process, and to analyze the resulting side effects such as parameter perturbations and optimization instability caused by explicit or implicit factual contradictions among training samples. We further construct \textbf{CFT-NoConf} and provide its construction pipeline, which can serve as a simple yet informative baseline to support future research.

\item \textbf{Hallucination Spillover Assessment.} 
We define the \textbf{Hallucination Cost} ($\mathcal{C}_h = \Delta_h / \Delta_f$) to measure how counterfactual training compromises the model's factual reliability outside the unlearning domain. Analogous to the Retain Cost, this metric is implemented by incorporating hallucination-related performance on the HaluEval dataset, enabling a principled assessment of whether a given method suffers from hallucination spillover.
\end{itemize}


\section{Implementation Details}
\label{apd:imple-details}

In this section, we outline several key details of our practical implementation.

\paragraph{Counterfactual target generation.} 
Following the sample construction methodology of the original RWKU benchmark, we directly invoke the LLaMA-3 model that serves as the unlearning target to generate the corresponding counterfactual samples, and apply the pipeline described in the main text to ensure that the generated samples are free of conflicts. We empirically observe that such samples yield notable improvements on gradient-based metrics, which validates the quality of the sample generation process.

\paragraph{Warmup.}
Following prior work on gradient analysis \citep{zhang2025resolvingeditingunlearningconflictsknowledge}, we incorporate an additional warmup phase before conducting our gradient diagnostics. This is motivated by the fact that unlearning methods are implemented on top of LoRA, which is conventionally initialized as an identity adapter ($B = 0$); computing $g_\theta$ at a freshly initialized LoRA adapter is therefore ill-defined. To remain consistent with established practice, we collect gradients and perform the corresponding evaluations only after the optimization has proceeded for a period of time.

Starting from $\theta^{(0)}$, we take $T = 100$ AdamW \citep{kingma2017adam} steps on a dedicated warmup subset $\mathcal{D}_s^{\text{cf,warm}}$ at learning rate $\eta = 5\times 10^{-5}$, yielding $\theta^{(T)}_s$, and compute all reported gradients at $\theta^{(T)}_s$. The held-out split ensures that none of the gradient-evaluation samples were seen during warmup. Empirically, we find that removing the warmup phase causes the structural signature of any input sample to collapse to $\mu \approx 0$, validating that warmup is necessary for the diagnostic to be meaningful.

\section{Detailed Related Work}
\label{apd:detailed_related_work}

\paragraph{Gradient Conflict in LLM Knowledge} represents a significant challenge in unlearning \citep{xu2025unlearning,wang2025uipeenhancingllmunlearning} and knowledge editing \citep{zhang-etal-2024-knowledge-graph,zhou2026uncovering,zhang2026spectral} tasks, where gradients derived from samples or subtasks associated with conflicting knowledge frequently exhibit negative cosine similarity, resulting in mutual cancellation, unstable optimization trajectories, and excessive parameter perturbations \citep{zhang2025resolvingeditingunlearningconflictsknowledge,yu2023unlearning}. Existing work on unlearning has primarily focused on inter-task gradient conflicts: for instance, \citet{zhang2025resolvingeditingunlearningconflictsknowledge} observed that gradients from knowledge updating and forgetting tasks tend to oppose each other, and proposed LOKA, which alleviates gradient antagonism by decoupling conflicting knowledge representations. More recent approaches such as GRU \citep{wang25gru} and LUR \citep{Patel_2025_ICCV_LUR} introduce gradient alignment correction techniques to balance the interference between forgetting and retention subtasks. These studies collectively demonstrate that naively aggregating conflicting gradients not only impedes convergence but also amplifies collateral damage to unrelated capabilities, which resonates with the knowledge conflict phenomenon we diagnose through the distribution of inter-sample gradient similarity. Building upon this line of research, our paper further investigates inter-sample gradient conflict scenarios and identifies corresponding issues within the emerging paradigm of counterfactual tuning.

\paragraph{Over-Generalization of Behavioral Biases} has been identified as a common side effect of post-training paradigms in prior work \citep{xue2026deactivating,shi-etal-2024-overkill,dabas2025justshiftsmitigatingoverrefusal}. In particular, refusal-based training—including its applications in safety alignment and the unlearning methods discussed in this paper—trains models to produce explicit refusal responses on target queries, and frequently leads to over-refusal on unrelated benign inputs due to the generalization of refusal templates \citep{shi2024muse,thaker2024guardrail}. Similarly, generalization behaviors associated with hallucination have been discussed in prior research: for example, \citet{lin-etal-2022-truthfulqa} examines how models produce generalizable false outputs by imitating common misconceptions present in training data, while \citet{Ji2023hallusurvey} discuss how hallucination biases acquired during training can generalize to unseen domains. Nevertheless, analogous issues in the unlearning setting have received limited attention. Our work identifies and systematically analyzes such over-generalization phenomena within the unlearning paradigm, offering empirical insights that we hope will inform and motivate future research in this direction.

\end{document}